\documentclass[conference]{IEEEtran}
\IEEEoverridecommandlockouts
\usepackage{cite}
\usepackage{amsmath,amssymb,amsfonts}
\usepackage{algorithm}
\usepackage{algorithmic}
\usepackage{graphicx}
\usepackage{textcomp}
\usepackage{xcolor}
\usepackage[utf8]{inputenc}
\usepackage{orcidlink}
\usepackage{stfloats}
\hypersetup{hidelinks}
\begin{document}

\title{FATHOMS-RAG: A Framework for the Assessment of Thinking and Observation in Multimodal Systems that use Retrieval Augmented Generation
\thanks{
Research sponsored by the Laboratory Directed Research and Development Program of Oak Ridge National Laboratory, managed by UT-Battelle, LLC, for the U. S. Department of Energy.

Notice: This manuscript has been authored by UT-Battelle, LLC under Contract No. DE-AC05-00OR22725 with the U.S. Department of Energy. The United States Government retains and the publisher, by accepting the article for publication, acknowledges that the United States Government retains a non-exclusive, paid-up, irrevocable, world-wide license to publish or reproduce the published form of this manuscript, or allow others to do so, for United States Government purposes. The Department of Energy will provide public access to these results of federally sponsored research in accordance with the DOE Public Access Plan (http://energy.gov/downloads/doe-public-access-plan).}
}

\author{
\IEEEauthorblockN{Samuel Hildebrand \orcidlink{0009-0004-6596-3104}}
\IEEEauthorblockA{\textit{Louisiana State University}\\
Baton Rouge, LA, USA \\
shilde2@lsu.edu}
\and
\IEEEauthorblockN{Curtis Taylor \orcidlink{0000-0002-2270-3538}}
\IEEEauthorblockA{\textit{Oak Ridge National Lab} \\
Oak Ridge, TN, USA  \\
taylorcr@ornl.gov}
\and
\IEEEauthorblockN{Sean Oesch \orcidlink{0000-0002-6909-1022}}
\IEEEauthorblockA{\textit{Oak Ridge National Lab} \\
Oak Ridge, TN, USA \\
oeschts@ornl.gov}
\and
\IEEEauthorblockN{James M. Ghawaly Jr. \orcidlink{0000-0001-8826-8500}}
\IEEEauthorblockA{\textit{Louisiana State University} \\
Baton Rouge, LA, USA\\
jghawaly@lsu.edu}
\and
\IEEEauthorblockN{Amir Sadovnik \orcidlink{0000-0001-9011-7365}}
\IEEEauthorblockA{\textit{Oak Ridge National Lab} \\
Oak Ridge, TN, USA \\
sadovnika@ornl.gov}
\and
\IEEEauthorblockN{Ryan Shivers \orcidlink{0000-0003-3124-4035}}
\IEEEauthorblockA{\textit{Oak Ridge National Lab} \\
Oak Ridge, TN, USA \\
shiversrm@ornl.gov}
\and
\IEEEauthorblockN{Brandon Schreiber \orcidlink{0009-0009-0389-6336}}
\IEEEauthorblockA{\textit{Oak Ridge National Lab} \\
Oak Ridge, TN, USA \\
schreiberb@ornl.gov}
\and
\IEEEauthorblockN{Kevin Kurian \orcidlink{0009-0009-4580-0574}}
\IEEEauthorblockA{\textit{University of Florida} \\
Gainesville, FL, USA \\
kuriankevin@ufl.edu}
}
\maketitle

\begin{abstract}
Retrieval-augmented generation (RAG) has emerged as a promising paradigm for improving factual accuracy in large language models (LLMs). We introduce a benchmark designed to evaluate RAG pipelines as a whole, evaluating a pipeline's ability to ingest, retrieve, and reason about several modalities of information, differentiating it from existing benchmarks that focus on particular aspects such as retrieval. We present (1) a small, human-created dataset of 93 questions designed to evaluate a pipeline's ability to ingest textual data, tables, images, and data spread across these modalities in one or more documents; (2) a phrase-level recall metric for correctness; (3) a nearest-neighbor embedding classifier to identify potential pipeline hallucinations; (4) a comparative evaluation of 2 pipelines built with open-source retrieval mechanisms and 4 closed-source foundation models; and (5) a third-party human evaluation of the alignment of our correctness and hallucination metrics.
We find that closed-source pipelines significantly outperform open-source pipelines in both correctness and hallucination metrics, with wider performance gaps in questions relying on multimodal and cross-document information. Human evaluation of our metrics showed average agreement of 4.62 for correctness and 4.53 for hallucination detection on a 1-5 Likert scale (5 indicating "strongly agree").
\end{abstract}


\section{Introduction}
Large language models (LLMs) have demonstrated remarkable capabilities across a wide range of tasks, yet their tendency to hallucinate factual details remains a limitation in domains requiring high reliability. Retrieval-augmented generation (RAG) has emerged as a strategy to mitigate this issue by grounding model outputs in external documents. Evaluating RAG pipelines for multimodal documents such as scientific PDFs remains under-explored, however. Scientific documents, among other document types, often encode key information in tables, diagrams, and figures, requiring modality-aware retrieval and reasoning to achieve reliable performance.

Existing benchmarks provide only partial coverage of this space. General-purpose multimodal evaluations such as MMMU and NPHardEval4V probe reasoning across modalities such as diagrams, charts, tables, scientific figures, graph visualizations and spatial puzzles, while specialized resources like CRAG target generation quality under retrieval conditions. However, no existing benchmark comprehensively assesses end-to-end RAG pipelines across ingestion, retrieval, reasoning, and generation for multimodal documents. Equally underexplored is the distinction between \emph{abstentions}, where a model refuses to answer due to insufficient context, and \emph{hallucinations}, where incorrect information is presented as fact. Current hallucination detection approaches often rely on access to hidden states or token-level probabilities, limiting their applicability to pipeline-level evaluations.

In this work, we address these gaps by introducing a benchmark for multimodal RAG evaluation with three core contributions. First, we construct a curated dataset of 93 questions spanning five categories: text-only, tables, images, multimodal, and cross-document multimodal, derived from eight open-access scientific papers. Second, we propose an evaluation methodology that combines phrase-level recall for correctness with a nearest-neighbor embedding classifier to automatically separate abstentions from hallucinations. Third, we provide a comparative evaluation of open-source and closed-source pipelines: (1) LlamaIndex with text-only ingestion, (2) Docling with EasyOCR for OCR and table recovery, and (3) closed-source multimodal APIs including Claude Sonnet-4, Gemini-2.5 Flash, GPT-4.1, and GPT-4o.

Our results demonstrate three consistent findings. First, text-only pipelines perform reasonably on purely textual queries but fail dramatically when important information is encoded in tables or images. Second, incorporating OCR and layout-aware preprocessing significantly improves performance on image and cross-document queries, narrowing the gap with closed systems. Third, closed-source APIs substantially outperform open-source pipelines in both correctness and hallucination avoidance, yet even state-of-the-art models struggle to answer questions that require information encoded in text and tables or images from multiple documents. This highlights a persistent bottleneck in cross-document multimodal reasoning across systems and motivates further research into modality-aware retrieval and structured reasoning approaches.

By releasing our dataset and evaluation framework, we provide a lightweight, reproducible tool for systematically benchmarking multimodal RAG pipelines. We aim for this work to support progress toward trustworthy retrieval-augmented systems by enabling comparative analysis across models, modalities, and ingestion strategies.

\section{Related Work}
The evaluation of multimodal reasoning capabilities has evolved from broad diagnostic assessments to specialized benchmarks targeting specific aspects of cross-modal understanding and retrieval-augmented generation. Our work builds upon this foundation by introducing a comprehensive benchmark for evaluating a RAG pipelines as a whole.

\subsection{General-Purpose Multimodal Benchmarks}
Early efforts to evaluate reasoning capabilities across diverse tasks include the MMLU-Pro benchmark by Wang \emph{et al.}, which provides a robust and challenging framework for assessing broad knowledge and reasoning in language models \cite{NEURIPS2024_ad236edc}. Expanding to multimodal settings, Yue \emph{et al.} introduce MMMU, a comprehensive multimodal understanding suite aimed at expert-level evaluation across multiple academic disciplines \cite{yue2024mmmu}. Fan \emph{et al.} propose NPHardEval4V, which dynamically generates NP-hard combinatorial puzzles paired with multimodal contexts to probe model adaptability \cite{fan2024nphardeval4vdynamicreasoningbenchmark}.

\subsection{Specialized Multimodal Benchmarks}
Beyond general-purpose evaluations, several benchmarks target specific reasoning challenges. Kil \emph{et al.} present MLLM-CompBench, measuring reasoning performance of multimodal LLMs on tasks from visual question answering to diagram interpretation \cite{kil2024mllm}. Guo \emph{et al.} introduce R-Bench, emphasizing graduate-level, cross-disciplinary problems spanning mathematics, biology, and engineering \cite{guo2025rbenchgraduatelevelmultidisciplinarybenchmarks}. EMMA by Hao \emph{et al.} evaluates reasoning over annotated image-text pairs, focusing on commonsense and factual inference in visual contexts \cite{hao2025mllmsreasonmultimodalityemma}.

\subsection{RAG System Evaluation and Hallucination Detection}
Yang \emph{et al.} introduce the CRAG benchmark, which evaluates RAG systems' generation capabilities under various retrieval conditions. However, CRAG focuses primarily on how generation behaves given retrieved context, rather than evaluating the retrieval mechanisms themselves or providing end-to-end pipeline assessment \cite{NEURIPS2024_1435d2d0}. 

RAGAS (Retrieval Augmented Generation Assessment) provides a framework for evaluating individual RAG components through metrics such as context precision, context recall, and answer relevancy \cite{es2023ragas}. While useful for component-level analysis, RAGAS does allow for evaluation of RAG pipelines as a whole.

Detecting hallucinations in model outputs has been approached from several directions. Recent work includes probabilistic uncertainty estimation through semantic entropy (Farquhar et al.) and model-internal state analysis for real-time detection (Su et al.) \cite{farquhar2024detecting, su2024unsupervisedrealtimehallucinationdetection}. While effective, these methods often require access to token-level probabilities, hidden states, or repeated sampling—conditions that are impractical when evaluating retrieval pipelines as whole, or impossible in many cases when using closed-source RAG pipelines.

\subsection{Our Contribution}
In contrast, our approach uses a nearest-neighbor classification framework over sentence embeddings to separate \emph{statement-style} answers from \emph{abstention-style} answers. By embedding both annotated examples and model outputs into a shared semantic space, abstentions such as "I cannot answer from the given context" can be automatically differentiated from answers asserted as factual statements. We combine this classification with phrase-level recall correctness scoring to identify hallucinations. Any response presented as a statement (not classified as an abstention) but failing to achieve full recall against ground truth is labeled as a hallucination. This embedding-based method provides an alternative that avoids dependence on model internals or handcrafted lexical rules.

While existing benchmarks excel at evaluating individual components (such as multimodal reasoning capabilities or retrieval mechanism effectiveness) none provide comprehensive evaluation of complete RAG pipelines. Our work addresses this gap by introducing a benchmark that evaluates the entire pipeline from document ingestion and retrieval through reasoning and answer generation, using curated question-answer pairs over multimodal PDF collections. By utilizing a small, hand-curated dataset derived from real scientific literature, we enable fast, systematic assessment and comparison of retrieval pipelines.

\section{Methodology}

\subsection{Dataset Construction}
\label{sect:Dataset}
To construct our evaluation dataset, we curated 93 non-trivial questions and associated answers from a collection of 8 research papers in the fields of AI and machine learning.\cite{wang2023docllmlayoutawaregenerativelanguage, utkarsh2025physicsconstrainedflowmatchingsampling, oesch2024pathautonomouscyberdefense, yang2025chartmimicevaluatinglmmscrossmodal, wang2025enigmaevalbenchmarklongmultimodal, fan2024nphardeval4vdynamicreasoningbenchmark, zhmoginov2022hypertransformermodelgenerationsupervised, zhang2024multimodalchainofthoughtreasoninglanguage} To answer these questions requires referencing various modalities within the 8 papers. The questions are organized into 5 categories: 
\begin{itemize}
\item Text-Only: These questions can be answered using only textual information from one of the dataset documents.
\item Tables: These questions require structured text extraction by referencing data in a table in one of the dataset documents.
\item Images: These questions require information only available in images in one of the dataset documents.
\item Multimodal: These questions require referencing data encoded in two or more of the text, image, or table modalities in one of the dataset documents. For instance, a question could require information from the methodology section of a paper, as well as data from a table in the results section, or information from a diagram in another section of the same paper.
\item Cross-Document Multimodal: These questions require information from two or more modalities in two or more papers.
\end{itemize}

All of the papers selected are available on \url{https://arxiv.org} and licensed under CC-BY. The full dataset of questions and answers, as well as all code used in our evaluation, can be found at \url{https://github.com/Sam-Hildebrand/FATHOMS-RAG}.

\subsection{Retrieved Answer Scoring}
For each question, one or more acceptable answers are created by human examination of the documents in the dataset. In some cases, multiple individual phrases may be required for a complete answer. For instance, the phrases "red", "green", and "blue" can appear in any order and with any amount of explanatory text between them in response to a question asking about the three most commonly used additive primary colors, but a fully correct answer must list all three. This metric is referred to as phrase-level recall.

Formally, let $A = \{a_1, a_2, \dots, a_m\}$ be the set of acceptable answers, where each $a_i = \{p_{i1}, p_{i2}, \dots, p_{i n_i}\}$ is a set of $n_i$ required phrases. Let $P$ denote the predicted text. Define the indicator function
\[
\mathbf{1}_{p \in P} =
\begin{cases}
1 & \text{if phrase $p$ is found in $P$}, \\
0 & \text{otherwise}.
\end{cases}
\]

The phrase-level recall of the prediction against a particular acceptable answer $a_i$ is
\[
R(a_i, P) = \frac{1}{n_i} \sum_{j=1}^{n_i} \mathbf{1}_{p_{ij} \in P}.
\]

Since any acceptable variation is valid, the overall correctness score of the predicted answer is defined as
\[
\text{Correctness}(P) = \max_{i=1,\dots,m} \; R(a_i, P).
\]

This evaluation method is represented algorithmically in Algorithm~\ref{alg:answer_scoring}. To evaluate the correctness of a pipeline's answer to a question, we represent ground truth answers as an array of arrays where the outer array contains all acceptable variations to an answer, and the inner array contains one or more key phrases that are required to be in that variation of the answer. The correctness of an answer is represented as a float between $0.0$ and $1.0$. This metric ensures that partial correctness is rewarded proportionally to the fraction of required phrases present, while the maximum operator selects the acceptable variation that best matches the predicted answer.

\begin{algorithm}
\caption{Pipeline Predicted Answer Scoring Against Ground Truth}
\label{alg:answer_scoring}
\begin{algorithmic}[1]
\STATE \textbf{Input:} predicted text, acceptable answers
\STATE \textbf{Output:} best score

\STATE $\text{best} \gets 0$
\FOR{each acceptable answer in acceptable answers}
    \STATE $\text{matches} \gets 0$
    \FOR{each phrase in acceptable answer}
        \IF{phrase found in predicted text}
            \STATE $\text{matches} \gets \text{matches} + 1$
        \ENDIF
    \ENDFOR
    \STATE $\text{score} \gets \text{matches} / \text{length of acceptable answer}$ \COMMENT{Number of phrases in acceptable answer}
    \STATE $\text{best} \gets \max(\text{best}, \text{score})$
\ENDFOR
\STATE \textbf{return} best

\end{algorithmic}
\end{algorithm}

\subsection{Nearest-Neighbor Based Hallucination Detection}
An important and under-explored aspect of evaluating LLM-based information retrieval pipelines is the ability to differentiate between answers where the model explicitly abstains from answering (for example, "I'm sorry, but I cannot answer the question with the context provided"), and answers which are incorrect but presented in a way that sounds factual. For the sake of distinction, we refer to the first category as \emph{abstentions}, and the second category as \emph{hallucinations}.  

To automatically separate these two cases, we employ a nearest-neighbor classifier trained on a small set of manually annotated examples of \emph{statement}-style answers and \emph{abstention}-style answers. Each training example is encoded using a sentence embedding model and stored in an index. At inference time, a predicted answer is embedded and classified by retrieving its nearest neighbor in the embedding space, which returns either the label "statement" or "abstention."  

Formally, let $f: \mathcal{T} \rightarrow \mathbb{R}^d$ be a sentence embedding function that maps text $t \in \mathcal{T}$ to a normalized $d$-dimensional embedding vector. Let $\{(x_k, y_k)\}_{k=1}^N$ be the set of training examples, where $x_k$ is an answer text and $y_k \in \{\text{statement}, \text{abstention}\}$. Their embeddings are given by  
\[
z_k = \frac{f(x_k)}{\|f(x_k)\|}, \quad k = 1, \dots, N.
\]

For a predicted answer $p$, we compute its embedding $z_p = \tfrac{f(p)}{\|f(p)\|}$ and find its nearest neighbor in the training set under cosine distance:  
\[
k^* = \arg\min_k \; \bigl(1 - z_p^\top z_k \bigr).
\]
The classification is then  
\[
C(p) = y_{k^*} \in \{\text{statement}, \text{abstention}\}.
\]

This classification is combined with the correctness score $\text{Correctness}(p)$ defined in Section~\ref{alg:answer_scoring} to determine hallucination status. We define the hallucination indicator function $H(p)$ as
\[
H(p) =
\begin{cases}
1 & \text{if } C(p) = \text{statement} \text{ and } \text{Correctness}(p) \neq 1.0, \\
0 & \text{otherwise}.
\end{cases}
\]

Thus, a predicted answer is considered a hallucination if and only if it is presented as a factual statement but fails to achieve full phrase-level recall. Abstentions, even if incorrect, are not considered hallucinations. The entire hallucination detection procedure is summarized in Algorithm~\ref{alg:nn_hallucination_detection}.

\begin{algorithm}
\caption{Detecting Hallucinations Using Nearest Neighbor Classification}
\label{alg:nn_hallucination_detection}
\begin{algorithmic}[1]
\STATE \textbf{Input:} predicted answer: text $\text{pred\_answer}$, correctness score: $\text{score}$
\STATE \textbf{Output:} Boolean indicating if the answer is a hallucination ($\text{true}$ or $\text{false}$)
\vspace{0.5em}
\STATE \textbf{function} Classify($\text{text}$)
\STATE \hspace{0.5em} $\text{embedding} \leftarrow \text{SentenceTransformer}(\text{text})$
\STATE \hspace{0.5em} $\text{nearest\_idx} \leftarrow \text{NearestNeighbor}(\text{embedding})$
\STATE \hspace{0.5em} \textbf{return} $\text{labels}[\text{nearest\_idx}]$ \COMMENT{Returns "statement" or "abstention"}
\STATE \textbf{end function}
\vspace{0.5em}
\STATE \textbf{function} IsHallucinating($\text{pred\_answer}$, $\text{score}$)
\STATE \hspace{0.5em} $\text{classification} \leftarrow \text{Classify}(\text{pred\_answer})$
\IF{$\text{classification} = \text{"statement"}$ \textbf{and} $\text{score} \neq 1.0$}
    \STATE \hspace{0.5em} \textbf{return} \textbf{true} \COMMENT{Classified as hallucination}
\ELSE
    \STATE \hspace{0.5em} \textbf{return} \textbf{false} \COMMENT{Not a hallucination}
\ENDIF
\STATE \textbf{end function}
\end{algorithmic}
\end{algorithm}

\section{Evaluations}
\subsection{LlamaIndex Text-Only Pipeline}
For the first evaluation of our benchmark, we implemented a retrieval-augmented generation (RAG) pipeline using \texttt{LlamaIndex} together with Ollama-hosted large language models. The full Python implementation is provided in our GitHub repository, but here we summarize the important design decisions.

All documents were ingested from a specified directory of PDF files and processed into pages using \texttt{PyMuPDF}. Each page of text was segmented into overlapping chunks using the \texttt{SentenceSplitter} class. We used a chunk size of 1024 characters with an overlap of 200 characters. Embeddings were generated using the Ollama embedding model \texttt{nomic-embed-text}, and responses were generated by the candidate LLM under evaluation. At query time, the pipeline performs dense retrieval using cosine similarity over the vector index. The top $k=3$ most relevant chunks are retrieved for each query and passed into the LLM. We believe this architecture to be representative of a basic, text-only RAG pipeline, although it may be in no way optimal for this type of document. 

Notably, this evaluation setup is only able to retrieve textual content. Images, figures, and tables in the PDFs are ignored during ingestion, which contributes to the high hallucination rate observed for multimodal question categories (see Appendix~\ref{tab:model_performance}). Thus, this evaluation specifically reflects the performance of text-only RAG pipelines under these conditions. Retrieved text is passed directly to the model in the prompt, and no additional multimodal or structured data handling is included in this configuration.   

\begin{figure}
    \centering
    \includegraphics[width=1.0\linewidth]{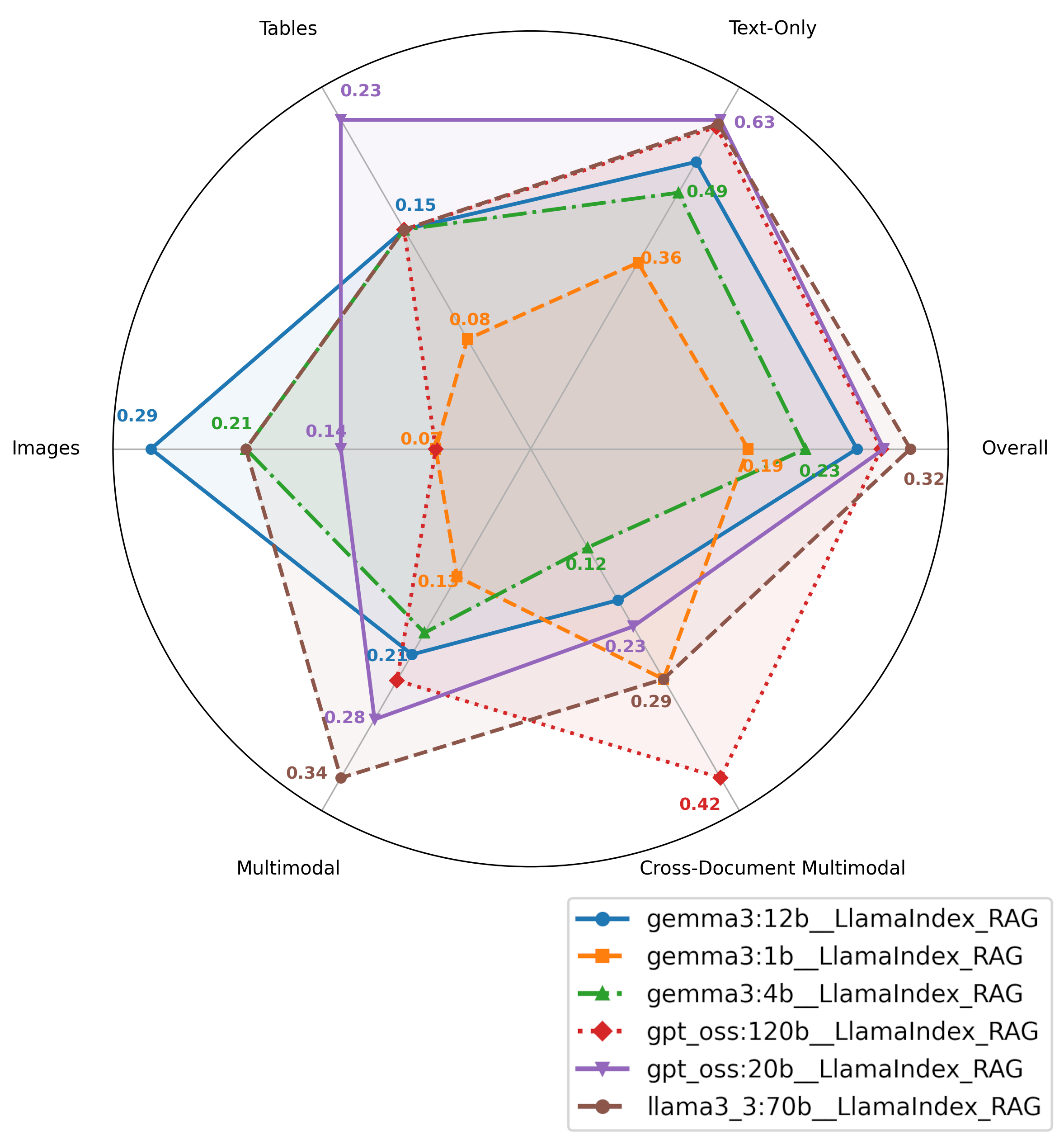}
    \caption{Answer Retrieval Accuracy for LlamaIndex Text Only RAG pipeline. (each axis normalized) Text-Only retrieval reaches 63\% accuracy, however, other categories recieve significantly lower scores.}
    \label{fig:LlamaIndex-Correctness}
\end{figure}

\begin{figure}
    \centering
    \includegraphics[width=1\linewidth]{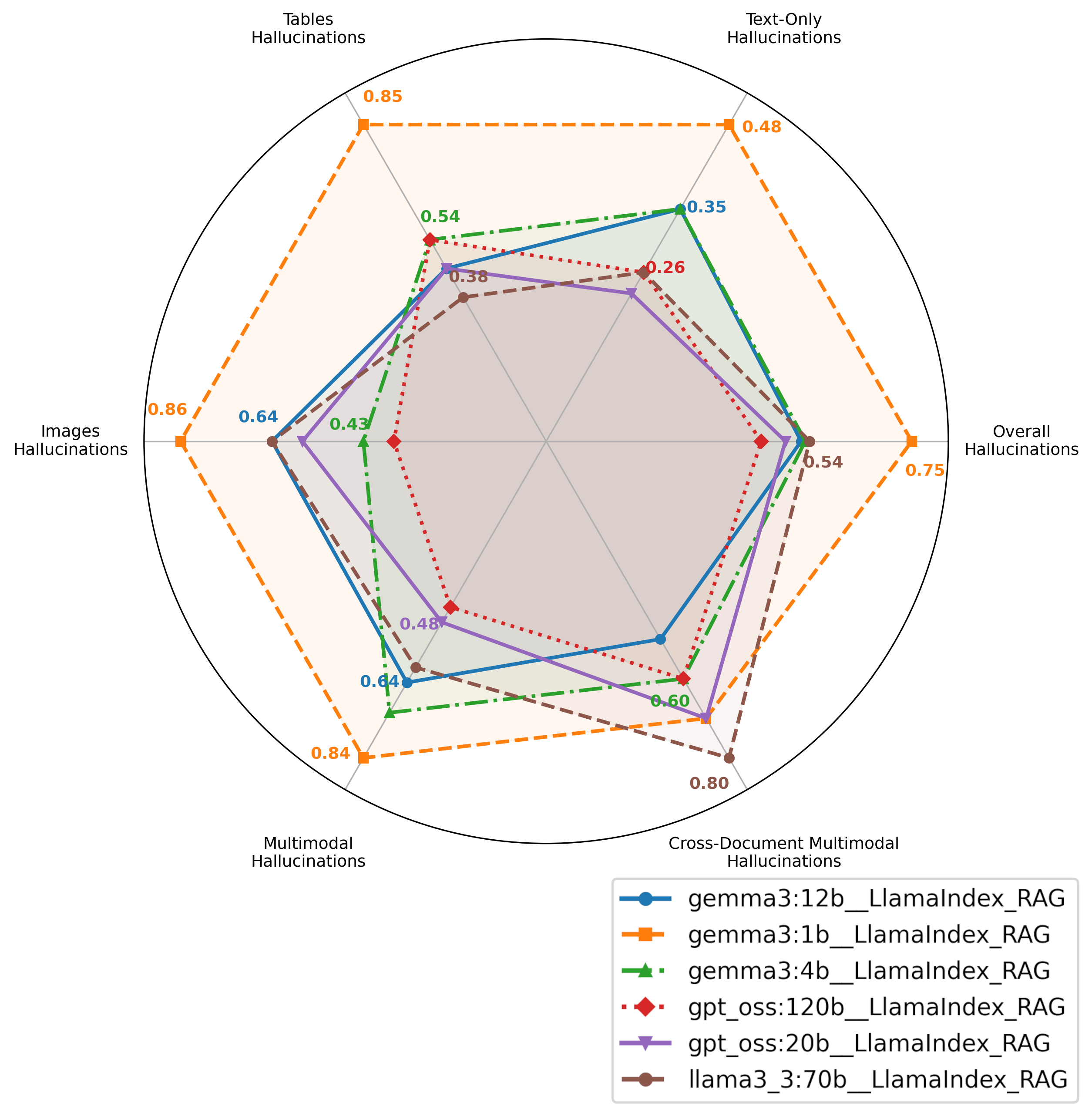}
    \caption{Calculated Hallucination Rate for LlamaIndex Text Only RAG pipeline. (Each axis normalized) Hallucinations reach as high as 86\% since this pipeline has no capability to process images.}
    \label{fig:LlamaIndex-Hallucinations}
\end{figure}

Because this pipeline cannot process non-textual content, queries requiring reasoning over tables, figures, or multimodal evidence are expected to perform poorly. Indeed, as shown in Figure~\ref{fig:LlamaIndex-Correctness} and Figure~\ref{fig:LlamaIndex-Hallucinations}, correctness scores for purely text-based queries are moderate, reaching up to 0.63 for the strongest mid-sized models (\texttt{gpt\_oss:20b}), and 0.62 for the largest (\texttt{llama3\_3:70b} and \texttt{gpt\_oss:120b}). However, performance drops sharply in other categories: table-based queries rarely exceed 0.23, and image-based queries often fall below 0.21. Multimodal correctness scores remain low (0.13–0.34), while hallucination rates in these categories regularly surpass 60–80\%. Note that for all correctness and hallucination figures we generate a radar graph with each axis normalized such that the highest score on each axis is placed on the outside perimeter to allow easier comparison in each question category.

Overall correctness across models ranges from 0.19 (\texttt{gemma3:1b}) to 0.32 (\texttt{llama3\_3:70b}), with hallucination ranging from 0.44 to 0.75. These results underscore two key limitations of the text-only design: (1) while larger models yield modest gains in recall on text-centric questions, their ability to avoid hallucination remains limited, and (2) the absence of modality-aware retrieval or encoding forces the system to respond to table and image questions with fabricated but confident-seeming text. Detailed numerical results for all models and categories are provided in Appendix~\ref{app:model_performance}.

\subsection{Docling with EasyOCR}
For the second evaluation of open-source pipelines, we implemented a RAG-style system using \texttt{Docling} with OCR enabled via \texttt{EasyOCR}. The pipeline converts PDF documents to markdown format while performing optical character recognition on all pages and extracting table structures with cell matching enabled. Documents are loaded at initialization, with content extracted and stored in memory. At query time, relevant documents are identified based on question metadata, and their full content (truncated to 15,000 characters if necessary) is injected directly into the prompt as context. The language model then generates answers based on this provided context. Default settings were used for both \texttt{Docling} and \texttt{EasyOCR}. Because this pipeline incorporates OCR and table-structure recovery, it is better suited to handle multimodal queries involving images or scanned text.

\begin{figure}
    \centering
    \includegraphics[width=1.0\linewidth]{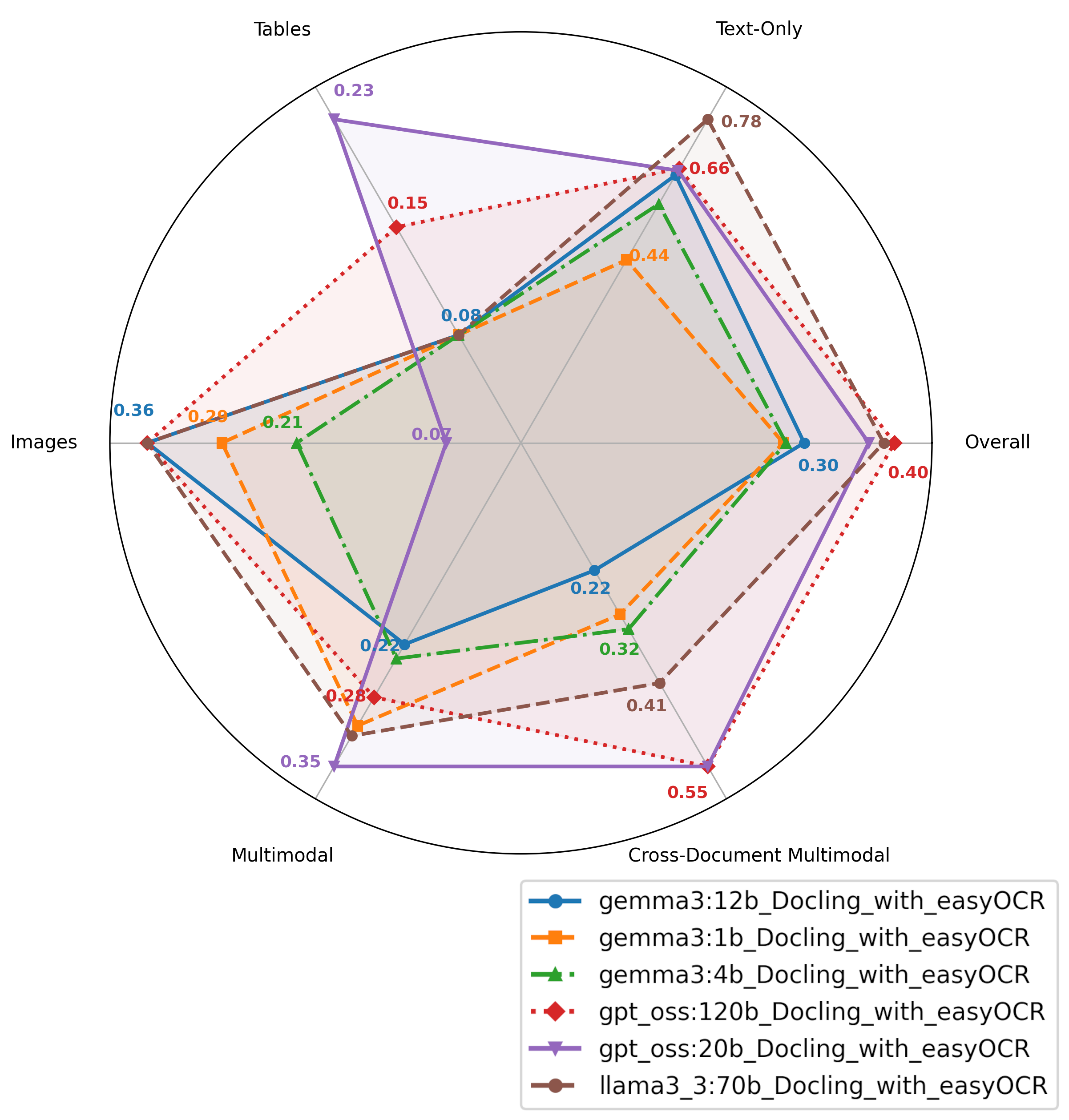}
    \caption{Answer Retrieval Accuracy for Docling and EasyOCR RAG pipeline. (Each axis normalized)}
    \label{fig:Docling-Correctness}
\end{figure}

\begin{figure}
    \centering
    \includegraphics[width=1\linewidth]{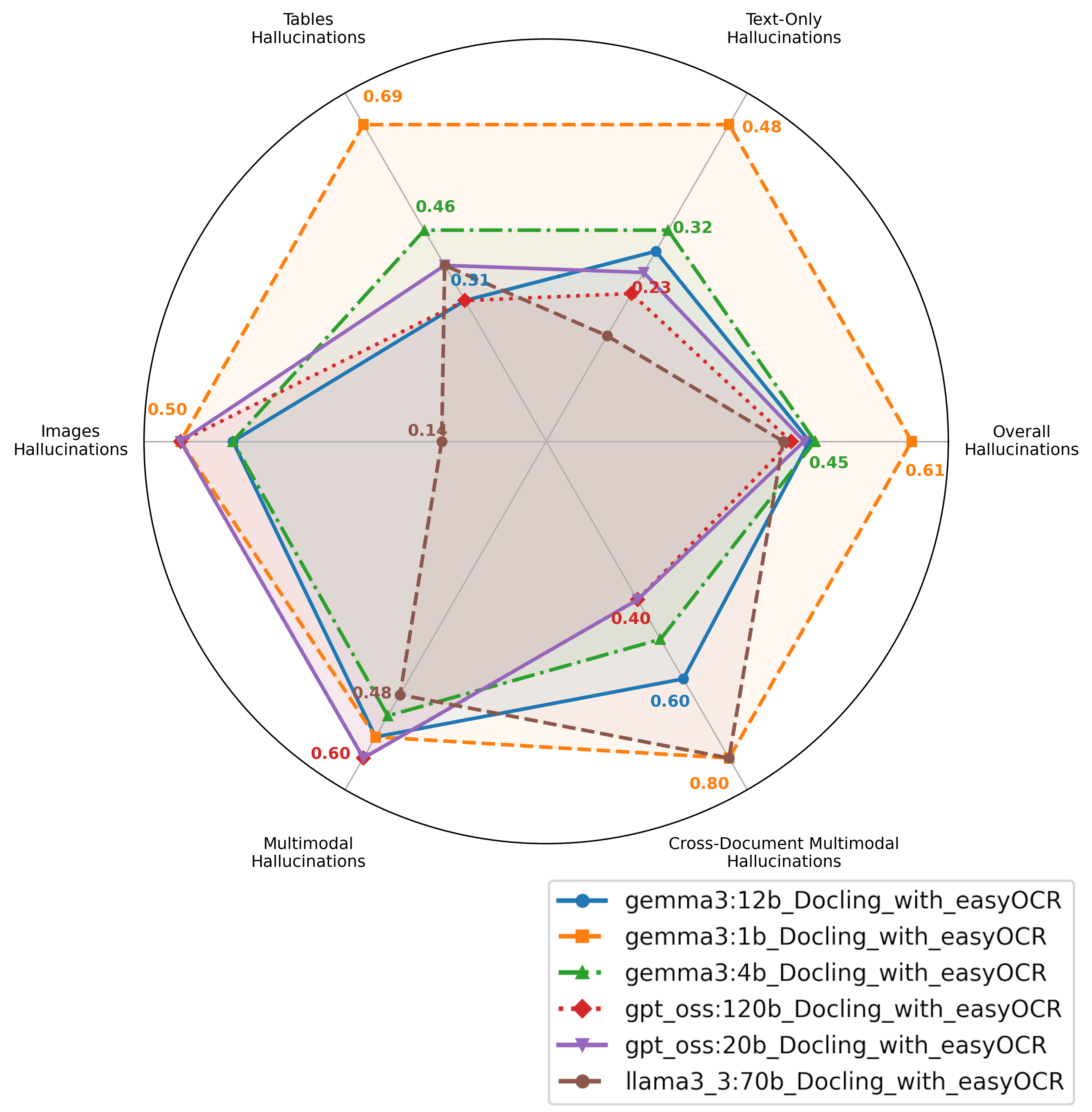}
    \caption{Calculated Hallucination Rate for Docling and EasyOCR RAG pipeline (Each axis normalized)}
    \label{fig:Docling-Hallucinations}
\end{figure}

As shown in Figure~\ref{fig:Docling-Correctness} and Figure~\ref{fig:Docling-Hallucinations}, text-only correctness scores reach as high as 0.78 (for \texttt{llama3\_3:70b}), with relatively low hallucination rates (down to 0.16). For table-based queries, performance remains weak (typically $\leq 0.23$), though hallucinations are somewhat reduced compared to the LlamaIndex pipeline. Image-based queries show notable improvements, with correctness up to 0.36 and hallucination rates as low as 0.14 for \texttt{llama3\_3:70b}. Cross-document multimodal questions see the largest gains: while LlamaIndex models rarely exceeded 0.29 correctness in this category, Docling and EasyOCR enabled strong models such as \texttt{gpt\_oss:120b} and \texttt{gpt\_oss:20b} to achieve correctness scores of 0.55, nearly doubling performance.

Overall correctness across models ranges from 0.28 (\texttt{gemma3:1b} and \texttt{gemma3:4b}) to 0.40 (\texttt{gpt\_oss:120b}), with hallucination rates spanning 0.39 to 0.61. These results highlight two key takeaways: (1) OCR-based document ingestion substantially improves coverage of image and scanned text queries and enables stronger cross-document reasoning, and (2) while hallucination rates are reduced compared to LlamaIndex, limitations in structured reasoning over tables and multimodal evidence still lead to frequent confident but incorrect responses. Detailed numerical results for all models and categories are provided in Appendix~\ref{app:model_performance}.

\subsection{Closed-Source Models}
In addition to the evaluation of open-source models and RAG pipelines, we also evaluated a set of closed-source API models including Claude Sonnet~4, Gemini~2.5 Flash, GPT-4.1, and GPT-4o. These models were accessed through their respective vendor APIs and integrated into the same evaluation harness described in Section~\ref{alg:answer_scoring}. Unlike the LlamaIndex pipeline, which operates strictly on text-only retrieval, these models natively support multimodal input and are capable of reasoning over embedded images and structured content such as tables.

Overall, we found the closed-source models to score substantially higher in correctness and with substantially fewer hallucinations compared to the open-source pipelines evaluated. Across all question categories, correctness scores range from $0.71$ to $0.82$, while hallucination rates are consistently lower, between 0.20 and 0.32 (see Appendix~\ref{app:model_performance} for full results). This stands in sharp contrast to the text-only pipeline, where correctness peaked at 0.33 and hallucination rates often exceeded $70\%$.

All of the closed-source API models we evaluated achieved very high preformance ($>80\%$ correct) on text-only and table-based queries, often exceeding $0.90$ correctness with hallucination rates below $0.10$. Image and multimodal categories remain challenging, but closed-source systems still show measurable robustness: correctness scores for image questions reach up to 0.86 with hallucination rates as low as 0.14. Finally, cross-document multimodal queries remain the hardest category, where correctness falls to 0.48--0.64 and hallucination rates rise as high as 0.60. These failure cases highlight the limits of current multimodal reasoning even in advanced proprietary systems.

\begin{figure}
    \centering
    \includegraphics[width=1\linewidth]{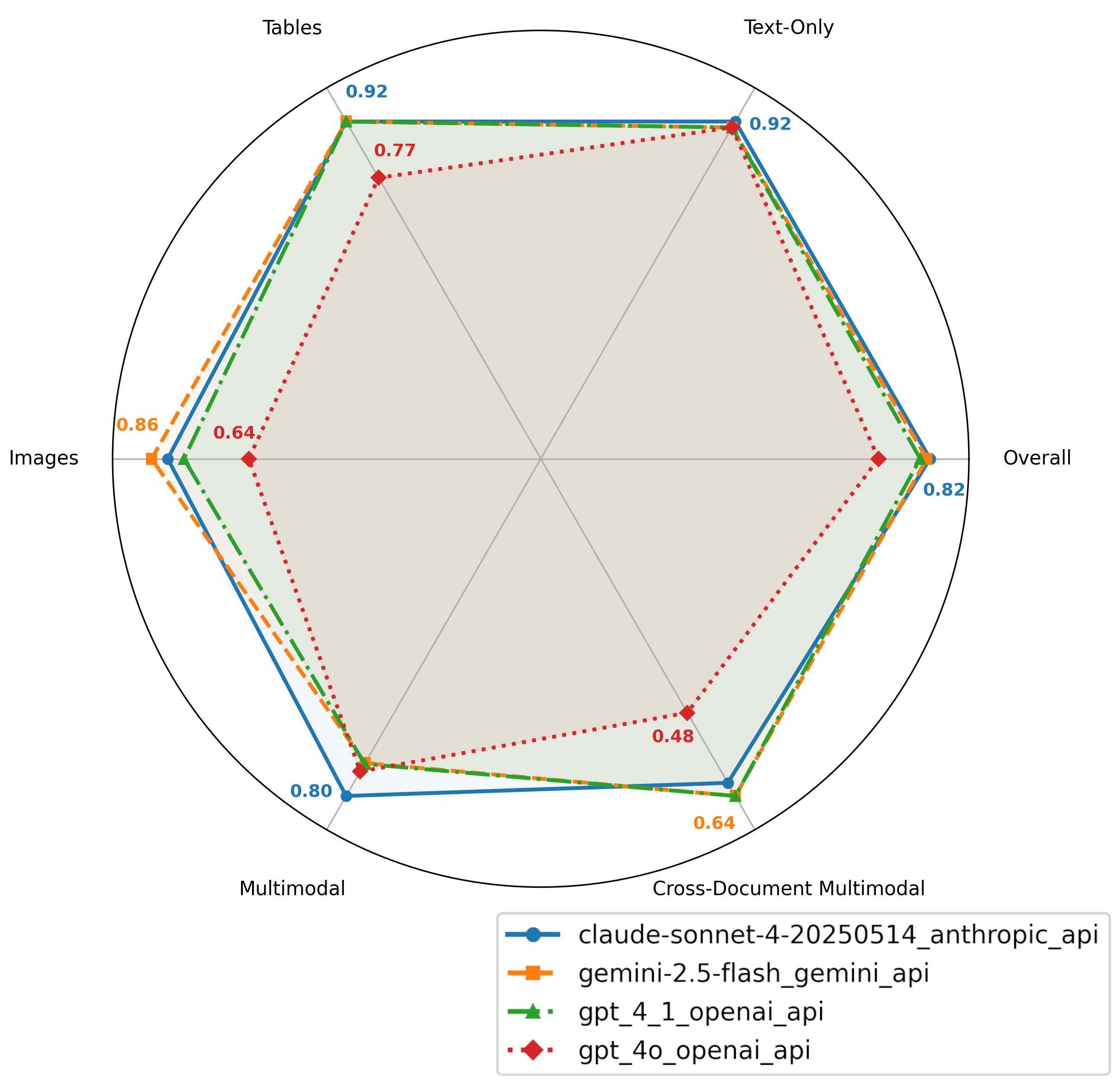}
    \caption{Answer Retrieval Accuracy for RAG pipelines of Closed-Source APIs. (Each axis normalized) Significantly higher scores in all categories.}
    \label{fig:Closed-Correctness}
\end{figure}

\begin{figure}
    \centering
    \includegraphics[width=1\linewidth]{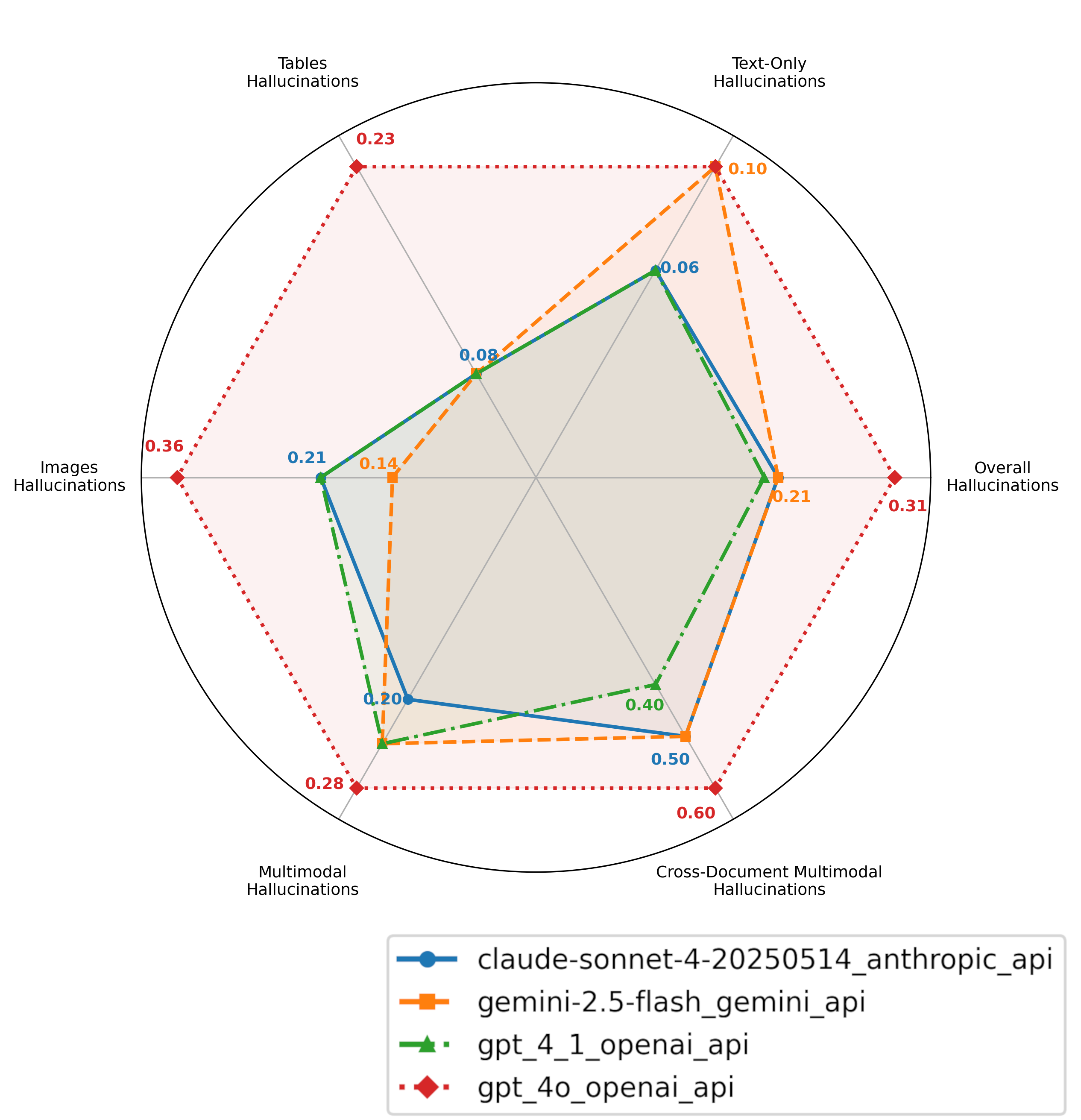}
    \caption{Calculated Hallucination Rate for RAG pipelines of Closed-Source APIs. (Each axis normalized) Notably lower hallucination rates in all categories except questions requiring reference to multiple documents (Cross-Document Multimodal).}
    \label{fig:Closed-Hallucinations}
\end{figure}

Figures~\ref{fig:Closed-Correctness} and \ref{fig:Closed-Hallucinations} visualize these results across categories, while Appendix~\ref{app:model_performance} provides detailed per-model tables. This evaluation indicates that closed-source APIs generate substantially fewer hallucination compared to text-only Retrieval Augmented Generation setups, but they still struggle in settings that demand cross-modal or cross-document reasoning.

\subsection{Human Evaluation of Correctness and Hallucination Metrics}

To validate the automatic scoring system used in our evaluations, we conducted a small-scale human study. 
We developed a script that randomly sampled 3 questions from each of the 16 evaluation result files, displaying the model prediction, ground truth, system-assigned correctness score, and suspected hallucination flag. 
A third-party reviewer was then asked to rate, on a five-point Likert scale, whether they agreed with the correctness and hallucination assignments. The reviewer was asked to rate their agreement on a scale from 1 (highly disagree) to 5 (highly agree) with a score of 3 being neutral. 
The results were aggregated by averaging across categories. Table~\ref{tab:human_eval} summarizes the human ratings of correctness and hallucination agreement.

\begin{table}[htb!]
\centering
\caption{Average human Likert ratings for correctness and hallucination detection. Ratings range from 1 (Strongly Disagree) to 5 (Strongly Agree).}
\label{tab:human_eval}
\resizebox{\columnwidth}{!}{%
\begin{tabular}{|l|c|c|}
\hline
\textbf{Category} & \textbf{Avg. Correctness Agreement} & \textbf{Avg. Hallucination Agreement} \\
\hline
Docling with EasyOCR & 5.00 & 4.89 \\
LlamaIndex RAG & 4.44 & 3.94 \\
Closed & 4.42 & 4.75 \\
\hline
Average & 4.62 & 4.53 \\
\hline
\end{tabular}%
}
\end{table}

We highlight here that overall the third-party human evaluation agreed with our correctness score with an average of 4.62 on the Likert scale and our hallucination score with an average of 4.53 on the Likert scale.

In reviewing our collected results from the human evaluation, we noted that phrases like "the provided excerpt does not contain" were not classified as an abstention 
by the current version of our hallucination classifier. Phrases such as these appeared twice in answers from LlamaIndex RAG category used in the human evaluation study, leading to a reduction in the hallucination agreement score for that category. We fully acknowledge that our current set of training data used in the sentence classifier for hallucination detection may not fully contain every possible variation of a statement that can be called an abstention, however we created an updated version of our statement / abstention training data for use in future evaluations.

\subsection{Comparison of Local RAG Pipelines with Closed-Source Models}
For text-only queries, LlamaIndex pipelines achieve moderate correctness, with scores up to 0.63 (\texttt{gpt\_oss:20b}) and 0.62 (\texttt{llama3\_3:70b} and \texttt{gpt\_oss:120b}), but hallucination rates remain elevated (0.23--0.48). Incorporating OCR and table recognition through Docling improves text-only correctness to as high as 0.78 (\texttt{llama3\_3:70b}) while reducing hallucination rates to 0.16--0.48. Image-based queries show the largest relative improvement under Docling, rising from 0.07--0.29 correctness in LlamaIndex to 0.29--0.36, with hallucinations dropping as low as 0.14. Cross-document multimodal queries also benefit: while LlamaIndex rarely exceeded 0.29 correctness, Docling enables strong models such as \texttt{gpt\_oss:120b} and \texttt{gpt\_oss:20b} to reach 0.55 correctness, nearly doubling performance. These findings underscore the importance of OCR and layout-aware preprocessing in mitigating the blind spots of text-only ingestion.

In contrast, closed-source models natively support multimodal reasoning and consistently outperform open-source pipelines. Claude Sonnet~4, Gemini~2.5 Flash, GPT-4.1, and GPT-4o all achieve overall correctness above 0.70 (peaking at 0.82 for Claude Sonnet~4) with substantially lower hallucination rates (0.20-0.32). Performance on text-only and table queries is near-perfect across all closed-source models ($\geq 0.90$ correctness, $\leq 0.10$ hallucinations). Image and multimodal queries remain more difficult, but closed systems still achieve correctness up to 0.86 with hallucination rates as low as 0.14, surpassing even the best OCR-based pipelines. Cross-document multimodal queries continue to be the hardest category, with correctness falling to 0.48-0.64 and hallucination rates reaching 0.40--0.60, yet these models still outperform both LlamaIndex and Docling setups.

Comparing the open-source evaluations with the closed API evaluations, we note: (1), OCR-based ingestion narrows, but does not eliminate the gap between local RAG pipelines and closed-source APIs. (2) Hallucination remains the primary challenge for open-source pipelines, particularly in multimodal settings, whereas closed systems exhibit greater robustness. (3) Cross-document multimodal reasoning is a universal bottleneck: while proprietary APIs perform better than local pipelines, all evaluated approaches struggle significantly in this category. Full per-model results are presented in Appendix~\ref{app:model_performance}.

\section{Conclusion}

We introduce a benchmark for evaluating retrieval-augmented generation (RAG) pipelines in multimodal contexts, with particular attention to hallucination detection and abstention classification. By combining phrase-level correctness scoring with a nearest-neighbor classifier for hallucination detection, our framework provides a lightweight and reproducible method for measuring the reliability of RAG systems across diverse document modalities.  

Our evaluations reveal three consistent findings. First, text-only pipelines such as LlamaIndex provide moderate performance on purely textual queries but fail dramatically when reasoning over tables, images, or multimodal evidence. Second, OCR- and layout-aware ingestion through Docling and EasyOCR narrows this gap by improving correctness on image and cross-document queries and reducing hallucination rates, though challenges with structured table reasoning persist. Third, closed-source models demonstrate substantially higher correctness and lower hallucination across nearly all categories, yet even state-of-the-art proprietary systems struggle with cross-document multimodal reasoning, highlighting this category as a universal bottleneck.  

As multimodal retrieval and reasoning become increasingly central to the deployment of trustworthy AI systems, evaluation frameworks must evolve to capture both correctness and reliability. Our benchmark represents a step toward this goal, underscoring the promise of open-source pipelines while highlighting the continued dominance and limitations of closed-source systems. Ultimately, progress in this area will hinge not only on scaling models, but also on developing principled methods for reducing hallucinations and enabling robust reasoning across modalities and documents.

\subsection{Limitations}
The primary limitation of this evaluation is the relatively constrained nature of our question dataset, both in size and scope. The small nature of the dataset allows for very fast evaluation of pipelines (~1 hour for most of the open-source pipelines), however for use cases outside of pipelines for scientific PDFs, this evaluation may not provide a useful benchmark. Future work will greatly expand the dataset with more questions and documents covering a larger subject area.

Our correctness metric, while interpretable, relies on exact phrase matching, which may underestimate partial semantic correctness. The nearest-neighbor hallucination classifier also depends on a small set of annotated examples and may misclassify borderline responses. Feedback from our human evaluation can be reimplemented into the hallucination detection model for better alignment with human expectations.

Additionally, comparison between open-source pipelines and closed-source APIs is constrained due to limited visibility into the retrieval mechanisms of closed-source models. Certain features like web search may play a role in compensating for poor retrieval ability in closed-source models that they obviously do not in open-source models. 

\subsection{Future Work}
Future work should greatly expand the size of the dataset used for evaluation. Such a dataset should focus on massively cross-document questions to test not only retrieval of a pipeline, but reasoning abilities of a model when given, say, 100 or more relevant chunks from various documents.

While our hallucination metric attempts to differentiate abstentions from statements of fact, it does not distinguishing between epistemic abstentions, that is, the pipeline abstains from answering due to insufficient information; and alignment-based abstentions, in which the model refuses to answer due to safety or ethical concerns. Future work could explore the different types of abstentions to gain deeper insight into model behavior when information is missing.

Additional future work might evaluate how tool call outside of a traditional RAG pipeline can influence the quality and correctness of the generated content. For instance, a pipeline used be researchers studying potential archaeological dig sites might be greatly improved qualitatively by calling a tool to generate a map of past dig sites.

\bibliographystyle{IEEEtran}
\bibliography{refs}

\appendices
\section{Full Table of Evaluation Results}
\label{app:model_performance}
All gathered correctness and hallucination scores are included in Table~\ref{tab:model_performance}. All of the code and raw result JSON files containing pipeline responses and scores can be found in our GitHub repository.

\begin{table*}[!]
\centering
\label{tab:model_performance}
\caption{Full List of Correctness and Hallucination Scores across all Evaluations}
\resizebox{\textwidth}{!}{%
\large
\begin{tabular}{|l|c|c|c|c|c|c|c|c|c|c|c|c|}
\hline
\textbf{Pipeline} & \textbf{Overall} & \textbf{Overall Hall.} & \textbf{Text-Only} & \textbf{Text-Only Hall.} & \textbf{Tables} & \textbf{Tables Hall.} & \textbf{Images} & \textbf{Images Hall.} & \textbf{MM} & \textbf{MM Hall.} & \textbf{Cross-Doc MM} & \textbf{Cross-Doc MM Hall.} \\
\hline
\multicolumn{13}{|c|}{\textbf{LlamaIndex Open Source Pipeline}} \\
\hline
gemma3:12b & 0.28 & 0.52 & 0.55 & 0.35 & 0.15 & 0.46 & 0.29 & 0.64 & 0.21 & 0.64 & 0.19 & 0.50 \\
\hline
gemma3:1b & 0.19 & 0.75 & 0.36 & 0.48 & 0.08 & 0.85 & 0.07 & 0.86 & 0.13 & 0.84 & 0.29 & 0.70 \\
\hline
gemma3:4b & 0.23 & 0.53 & 0.49 & 0.35 & 0.15 & 0.54 & 0.21 & 0.43 & 0.19 & 0.72 & 0.12 & 0.60 \\
\hline
gpt\_oss:120b & 0.30 & 0.44 & 0.62 & 0.26 & 0.15 & 0.54 & 0.07 & 0.36 & 0.24 & 0.44 & 0.42 & 0.60 \\
\hline
gpt\_oss:20b & 0.30 & 0.49 & 0.63 & 0.23 & 0.23 & 0.46 & 0.14 & 0.57 & 0.28 & 0.48 & 0.23 & 0.70 \\
\hline
llama3\_3:70b & 0.32 & 0.54 & 0.62 & 0.26 & 0.15 & 0.38 & 0.21 & 0.64 & 0.34 & 0.60 & 0.29 & 0.80 \\
\hline
\multicolumn{13}{|c|}{\textbf{Docling with EasyOCR Pipeline}} \\
\hline
gemma3:12b & 0.30 & 0.44 & 0.65 & 0.29 & 0.08 & 0.31 & 0.36 & 0.43 & 0.22 & 0.56 & 0.22 & 0.60 \\
\hline
gemma3:1b & 0.28 & 0.61 & 0.44 & 0.48 & 0.08 & 0.69 & 0.29 & 0.50 & 0.31 & 0.56 & 0.29 & 0.80 \\
\hline
gemma3:4b & 0.28 & 0.45 & 0.58 & 0.32 & 0.08 & 0.46 & 0.21 & 0.43 & 0.24 & 0.52 & 0.32 & 0.50 \\
\hline
gpt\_oss:120b & 0.40 & 0.41 & 0.66 & 0.23 & 0.15 & 0.31 & 0.36 & 0.50 & 0.28 & 0.60 & 0.55 & 0.40 \\
\hline
gpt\_oss:20b & 0.37 & 0.43 & 0.66 & 0.26 & 0.23 & 0.38 & 0.07 & 0.50 & 0.35 & 0.60 & 0.55 & 0.40 \\
\hline
llama3\_3:70b & 0.39 & 0.39 & 0.78 & 0.16 & 0.08 & 0.38 & 0.36 & 0.14 & 0.32 & 0.48 & 0.41 & 0.80 \\
\hline
\multicolumn{13}{|c|}{\textbf{Closed Source API Pipeline}} \\
\hline
claude-sonnet-4 & 0.82 & 0.21 & 0.92 & 0.06 & 0.92 & 0.08 & 0.82 & 0.21 & 0.80 & 0.20 & 0.62 & 0.50 \\
\hline
gemini-2.5-flash & 0.81 & 0.21 & 0.90 & 0.10 & 0.92 & 0.08 & 0.86 & 0.14 & 0.72 & 0.24 & 0.64 & 0.50 \\
\hline
gpt\_4\_1 & 0.79 & 0.20 & 0.90 & 0.06 & 0.92 & 0.08 & 0.79 & 0.21 & 0.72 & 0.24 & 0.64 & 0.40 \\
\hline
gpt\_4o & 0.71 & 0.31 & 0.90 & 0.10 & 0.77 & 0.23 & 0.64 & 0.36 & 0.74 & 0.28 & 0.48 & 0.60 \\
\hline
\end{tabular}%
}
\end{table*}

\end{document}